\documentclass[sigconf,nonacm]{acmart}

\AtBeginDocument{%
  }

\usepackage{multirow}
\usepackage{booktabs}
\usepackage{graphicx}
\usepackage{wrapfig}   % 用于包裹图片

\usepackage{amsthm}     % 定理环境支持


\usepackage{mathtools}  % amsmath的扩展

\usepackage{mathrsfs}   % 花体字符
\usepackage{bm}         % 粗体数学符号

\usepackage{geometry}   % 页面尺寸设置
\usepackage{xcolor}     % 颜色支持
\usepackage{tcolorbox}  % 彩色文本框
\usepackage{enumitem}   % 列表环境定制

\usepackage{graphicx}   % 插图
\usepackage{tikz}       % 绘图
\usepackage{subcaption}

\theoremstyle{plain}
\newtheorem{theorem}{Theorem} 
\newtheorem{lemma}{Lemma}   
 
\newtheorem{corollary}{Corollary}    
\theoremstyle{definition}
    
\theoremstyle{remark}
\newtheorem{remark}{Remark}

\tcbuselibrary{theorems}

\begin{document}

%%
%% The "title" command has an optional parameter,
%% allowing the author to define a "short title" to be used in page headers.
\title{Scaling Laws for Data-Efficient Visual Transfer Learning}

%%
%% The "author" command and its associated commands are used to define
%% the authors and their affiliations.
%% Of note is the shared affiliation of the first two authors, and the
%% "authornote" and "authornotemark" commands
%% used to denote shared contribution to the research.

\author{
    Wenxuan Yang$^{1}$, 
    Qingqu Wei$^{1}$, 
    Chenxi Ma$^{1}$, 
    Weimin Tan$^{1}$, 
    Bo Yan$^{1}$ \\
    $^1$Shanghai Key Laboratory of Intelligent Information Processing, School of Computer Science, Fudan University, Shanghai, China \\
}

\renewcommand{\shortauthors}{Yang et al.}

%%
%% The abstract is a short summary of the work to be presented in the
%% article.
\begin{abstract}
Current scaling laws for visual AI models focus predominantly on large-scale pretraining, leaving a critical gap in understanding how performance scales for ‌data-constrained downstream tasks‌. To address this limitation, this paper establishes the first practical framework for data-efficient scaling laws in visual transfer learning, addressing two fundamental questions: \textbf{1) How do scaling behaviors shift when downstream tasks operate with limited data?} \textbf{2) What governs the efficacy of knowledge distillation under such constraints?} Through systematic analysis of vision tasks across data regimes (1K–1M samples), we propose the distillation boundary theory, revealing a critical turning point in distillation efficiency: 1) Distillation superiority: In data-scarce conditions, distilled models significantly outperform their non-distillation counterparts, efficiently leveraging inherited knowledge to compensate for limited training samples.
2) Pre-training dominance: As pre-training data increases beyond a critical threshold, non-distilled models gradually surpass distilled versions, suggesting diminishing returns from knowledge inheritance when sufficient task-specific data becomes available. Empirical validation across various model scales (2.5M to 38M parameters) and data volumes demonstrate these performance inflection points, with error difference curves transitioning from positive to negative values at critical data thresholds, confirming our theoretical predictions. This work redefines scaling laws for data-limited regimes, bridging the knowledge gap between large-scale pretraining and practical downstream adaptation, addressing a critical barrier to understanding vision model scaling behaviors and optimizing computational resource allocation.
\end{abstract}

% \begin{CCSXML}
% <ccs2012>
%    <concept>
%        <concept_id>10010147.10010178.10010224</concept_id>
%        <concept_desc>Computing methodologies~Computer vision</concept_desc>
%        <concept_significance>500</concept_significance>
%        </concept>
%  </ccs2012>
% \end{CCSXML}

% \ccsdesc[500]{Computing methodologies~Computer vision}

%  \keywords{Scaling Laws, Transfer Learning, Foundation Models}

\maketitle

\section{Introduction}
\begin{figure}[t!]
    \centering
    \begin{minipage}[b]{0.9\linewidth}
    \includegraphics[width=1.0\textwidth]{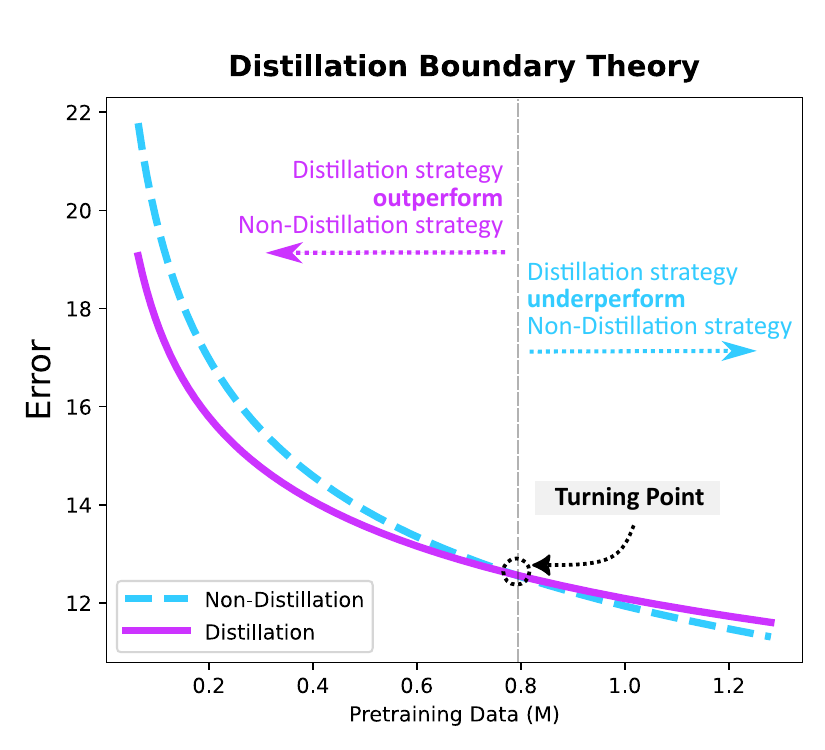}
    \end{minipage}
    \vspace{-15pt}
    \caption{Visualization of the Distillation Boundary Theory. Empirical results reveal the existence of a critical pretraining data threshold: prior to this threshold, distillation enhances model generalization, while beyond it, the added constraints of distillation may hinder the model's capacity to fully benefit from abundant data, leading to a decline in performance.}
    \vspace{-5pt}
    \label{fig:head-intersection}
\end{figure}
In recent years, transformer-based deep learning models have revolutionized multiple fields \cite{transformer}, demonstrating remarkable performance, particularly in natural language processing (NLP) \cite{nlptransformer1,nlptransformer2} and computer vision (CV)\cite{visiontransformer1,visiontransformer2}. Researchers have thoroughly investigated the relationship between model size, training data volume, and performance outcomes, discovering a fundamental principle: When computational resources are sufficient, both data scale and model parameters should increase according to power-law growth to maximize performance \cite{powerlaw1,powerlaw2}. This insight establishes the foundation of scaling laws and provides theoretical guidance for large-scale pretraining approaches. In NLP, OpenAI's GPT series\cite{gpt1,gpt2,gpt3,gpt4} has demonstrated how simultaneously increasing model parameters and training data significantly enhances generation and understanding capabilities. The CV field has seen similar validation through the Vision Transformer (ViT) \cite{vit}, where models like ViT-L/16 with 320 million parameters, trained on the extensive JFT-300M dataset \cite{jft300m}, achieved substantial improvements in visual tasks. However, despite confirming the effectiveness of scaling laws under conditions of large-scale data and abundant computational resources, current research predominantly focuses on pretraining scenarios, leaving a significant knowledge gap in understanding scaling behaviors for downstream applications, particularly under data-limited conditions.

Therefore, analyzing scaling laws under data-limited conditions is particularly significant because in practical research and applications, we frequently encounter the challenge of limited data resources. Existing public datasets (such as ImageNet \cite{imagenet}, Open Images \cite{openimages}, COCO\cite{coco}) are relatively limited in scale (<20M), while larger datasets (>100M, such as JFT-300M \cite{jft300m}, ALIGN 1.8B\cite{align}, JFT-3B\cite{jft3B}, IG-1B-Targeted\cite{ig}) are typically not publicly available. Therefore, compared to studying scaling laws under large-scale data conditions, a more pressing and practically relevant question is: \textbf{How do scaling behaviors shift when downstream tasks operate with limited data?} This question is particularly critical in the field of transfer learning, as pre-trained models typically require fine-tuning for specific downstream applications, and the significant disparities in data across research domains\cite{cvdataset} make the transfer process from general pretraining to specific application scenarios exceptionally complex. However, the intrinsic relationship between model size, pretraining data volume, and downstream task data volume has not been systematically investigated, especially in data-constrained environments (1K-1M samples) \cite{dataconsl}. Notably, current research severely lacks a theoretical analysis framework for the complex interaction mechanisms among these three key factors. This knowledge gap not only hinders our deeper understanding of the scaling behaviors of large vision models but also significantly limits the possibilities for developing model optimization strategies and efficient computational resource allocation, making it an urgent issue to address.

To address this critical limitation, we establish the first practical framework for data-efficient scaling laws in visual transfer learning. Through systematic analysis of vision tasks across four benchmarks, we propose a comprehensive scaling law that reveals the power-law relationships between model size, pretraining data, and fine-tuning data volumes. This practical framework provides a novel foundation for visual transfer learning, enabling researchers to predict and optimize visual model performance more accurately in data-constrained scenarios.

Building on our proposed scaling law, we further explore a second fundamental question: \textbf{What governs the efficacy of knowledge distillation under such constraints?} Through theoretical analysis and experimental validation across various model scales (2.5M to 38M parameters), we propose the distillation boundary theory, revealing a phase transition in distillation efficiency. We identify a critical inflection point where distilled models significantly outperform non-distilled ones in data-scarce conditions by leveraging inherited knowledge; however, as pretraining data exceeds a specific threshold, non-distilled models gradually surpass distilled versions as knowledge inheritance benefits diminish. Our empirical validation demonstrates these performance inflection points, with error difference curves transitioning from positive to negative values at critical data thresholds, confirming our theoretical predictions.

The core contributions of this paper are as follows:
\begin{itemize}
    \item This paper first establish the systematic scaling law for visual downstream tasks under data-efficient conditions, revealing power-law relationships between model size, pretraining data, and fine-tuning data volumes. This provides a theoretical foundation for performance prediction in data-constrained scenarios.
    \item We quantify how pretraining data size affects downstream performance, highlighting domain consistency importance and revealing diminishing returns as data scales increase.
    \item Through theoretical and experimental work, we propose the distillation boundary theory for model reusing tasks, identifying a critical inflection point where distilled models initially outperform non-distilled models as fine-tuning data increases but are subsequently surpassed beyond a threshold. This theory guides resource allocation and performance optimization.
    % \item We introduce the comprehensive research on scaling laws for visual knowledge distillation in data-constrained environments, showing it follows power-law relationships while substantially improving smaller model performance with minimal data requirements.
\end{itemize}
\section{RELATED WORK}
As the introduction mentions, transformer architecture has become the foundational building block of modern deep learning models in various areas. In recent years, extensive research has been dedicated to understanding the scaling laws that govern the relationship between model size, dataset size, and computational resources. These scaling laws provide critical insights into optimizing training efficiency across various tasks, including NLP, multimodal learning, and downstream training, enabling researchers to reduce computational costs while maintaining strong model performance. This section reviews key contributions related to such areas.

\subsection{Scaling Laws in NLP}
Early studies \cite{hestness2017deep,tan2019efficientnet,gpt3} explore the relationship between model performance and scale. Although the concept of scaling laws is not initially formalized, these pioneering works provide crucial insights and lay the groundwork for understanding the predictability of model performance as scale increases.

In NLP, the concept of scaling laws is widely recognized largely due to the works of \cite{kaplan2020scaling, henighan2020scaling, hernandez2021scaling}. These studies formally establish a quantitative relationship between model size, dataset size, and training compute for transformer models. Their findings have not only provided a foundation for subsequent research, but have also inspired deeper investigations into the fundamental principles governing scaling behavior in language models.

Beyond empirical exploration of neural scaling laws across different models and tasks, many researchers have sought to challenge these patterns by refining model training strategies or data sampling approaches in hopes of breaking conventional scaling laws. \cite{wang2023data} proposes a novel approach to *breaking* the neural scaling law by initializing large models with the weights of smaller pretrained models, thus improving data efficiency. \cite{tay2021scale} has further shown that model shape, rather than model size, plays a more crucial role in downstream NLP tasks and has introduced a scaling strategy called DeepNarrow, which reduces parameter count while improving training speed and performance.

\subsection{Scaling Laws in  Multimodal Learning} 

Apart from the scaling laws of transformer models in NLP, various other domains have also been progressively explored. The scaling behavior of mixture-of-experts (MoE) models with trillions of parameters has been analyzed in \cite{fedus2022switch}. A comprehensive examination of scaling laws has been conducted in \cite{tay2022scaling}.

The scaling dynamics of Vision Transformers (ViTs) and data, both in expansion and reduction, have been investigated in \cite{jft3B}, which establishes relationships between error rate, data availability, and compute resources. \citet{alabdulmohsin2023getting} further examines how scaling laws can be leveraged to optimize the design of Vision Transformer architectures, enabling them to outperform larger models of the same type.

Similarly to \cite{ wang2023data, tay2021scale}, \cite{sorscher2022beyond} challenges traditional scaling laws in the CV area by introducing a simple yet effective data pruning method, demonstrating that careful data selection can significantly reduce computational costs without compromising performance. 

\subsection{ Scaling Laws for Downstream Tasks}

Previous studies have primarily focused on exploring the relationship between large models and large datasets within upstream tasks without considering their applicability to downstream tasks. Recently, however, several works have investigated scaling laws in the context of downstream applications.

\citet{hernandez2021scaling} analyzes scaling laws in both pretraining and fine-tuning, introducing the concept of effective data transferred, which quantifies the efficiency of pretraining data in downstream performance. Additionally, \citet{geiping2023cramming} examines the downstream capabilities of a Transformer-based language model trained entirely from scratch within a single day on a single GPU, leveraging scaling laws as a framework for evaluation. Furthermore, \cite{prato2021scaling, aghajanyan2023scaling,yang2024optimizing} have identified that in compute-optimal training, model size and the number of training tokens should be scaled proportionally. More recently, studies such as \cite{minderer2023scaling, tschannen2023image,yang2024medical} have aimed to enhance downstream task performance by expanding the size of training datasets.

\section{Scaling Laws for Transfer Learning}
\subsection{Data and Model Size Scaling Methodology}

\vspace{2pt}
\noindent
\textbf{Data size variation.} To systematically study the impact of data scale on model performance, we employ uniform sampling methods for both upstream and downstream datasets. While maintaining the same number of categories, we sample 5\%, 10\%, 25\%, 33\%, 50\%, 70\%, and 100\% of the examples from both pre-training and fine-tuning datasets. This granular sampling approach allows us to precisely capture the relationships between data volume and model performance at different scales. By systematically varying both upstream and downstream data quantities, we can comprehensively fit the scaling laws specifically tailored for downstream tasks. This methodology enables us to quantify how pre-training data efficiency translates to downstream task performance and identify optimal data allocation strategies.

\vspace{2pt}
\noindent
\textbf{Model size variation.} We maintain a fixed size of 64 for each attention head in the Deit \cite{touvron2021training} model, and vary the number of heads to adjust the overall model size. Specifically, we implement configurations with 2, 4, 6, and 8 attention heads. This approach is based on the finding from \cite{kaplan2020scaling} that compared to the total number of parameters, the specific shape of the model has negligible influence on the performance of large transformers. This methodology allows us to systematically investigate how model capacity affects performance across different data regimes.

\subsection{A Scaling Law for Downstream Performance}
\label{power_law}
Similar to cross-entropy and perplexity which follow power-law scaling behavior, we find that both error rate and cross-entropy loss for downstream tasks also exhibit predictable scaling properties. While previous research has established scaling laws for upstream pretraining performance \cite{kaplan2020scaling,hernandez2021scaling,henighan2020scaling,jft3B,wang2023data, busbridge2025distillation}, systematic studies on comprehensive scaling laws for downstream task performance remain limited, particularly when considering the combined effects of pretraining data size, model size, and fine-tuning data size.

Through experimental analysis across multiple vision downstream tasks, we propose two related downstream scaling laws:

\begin{itemize}
    \item \textbf{For error rate:} 
    \begin{equation}
    E(D_p, M, D_f) = E_{\infty} + \frac{D_p^{-\alpha}}{\lambda_p} + \frac{M^{-\beta}}{\lambda_m} + \frac{D_f^{-\gamma}}{\lambda_f} \label{e_error}
    \end{equation}
    
    \item \textbf{For cross-entropy loss:} 
    \begin{equation}
    L(D_p, M, D_f) = L_{\infty} + \frac{D_p^{-\alpha''}}{\omega_p} + \frac{M^{-\beta''}}{\omega_m} + \frac{D_f^{-\gamma''}}{\omega_f} \label{e_loss}
    \end{equation}
\end{itemize}

Here, $D_p$ represents pretraining data size, $M$ denotes model size, and $D_f$ indicates fine-tuning data size. The parameters $E_{\infty}$ and $L_{\infty}$ represent the irreducible error and loss, respectively. The power-law exponents $\alpha$, $\beta$, $\gamma$, $\alpha'$, $\beta'$, and $\gamma'$ govern how quickly performance improves as each scaling factor increases. The scaling parameters $\lambda_p$, $\lambda_m$, $\lambda_f$, $\omega_p$, $\omega_m$, and $\omega_f$ determine the characteristic scale of each dimension.

Through our experimental analysis (from Table ~\ref{tab:scaling_coefficients}) across multiple vision downstream datasets, we find that pretraining data scale $(D_p)$ exerts the most profound influence on downstream task performance, followed by model scale $(M)$, while fine-tuning data quantity $(D_f)$ has a comparatively modest effect. This provides practical guidance for model development under resource constraints: increasing pretraining data typically yields more effective improvements compared to expanding model size or increasing fine-tuning data.

However, in practical applications, acquiring large-scale pretraining data often faces numerous challenges: high costs, data privacy issues, and scarcity of domain-specific data. This raises a key question: \textbf{When additional pretraining data is unavailable, how can we maximize the efficiency of limited resources?}

\begin{figure*}[h!]
  \centering
  \begin{minipage}[b]{0.9\linewidth}
    \includegraphics[width=\linewidth]{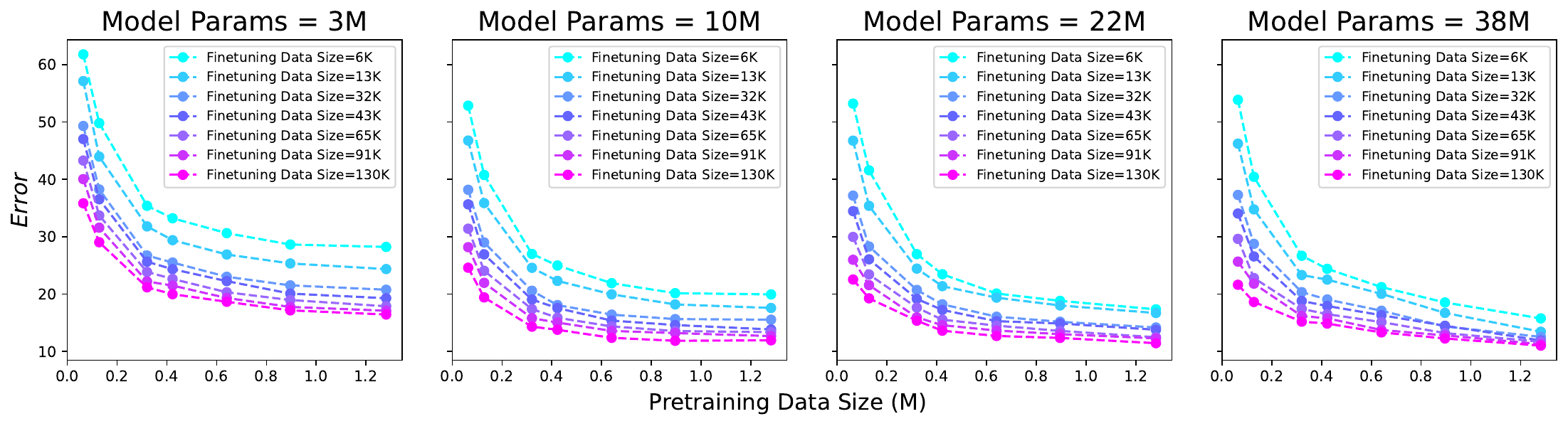}
    \vspace{-10pt}
    \subcaption{Scaling laws for pre-training data size on ImageNet100. Illustration of the impact of pre-training data volume ($Dp$) on error rate across different model sizes ($M$) on ImageNet100.}
  \end{minipage}

\begin{minipage}[b]{0.9\linewidth}
    \includegraphics[width=\linewidth]{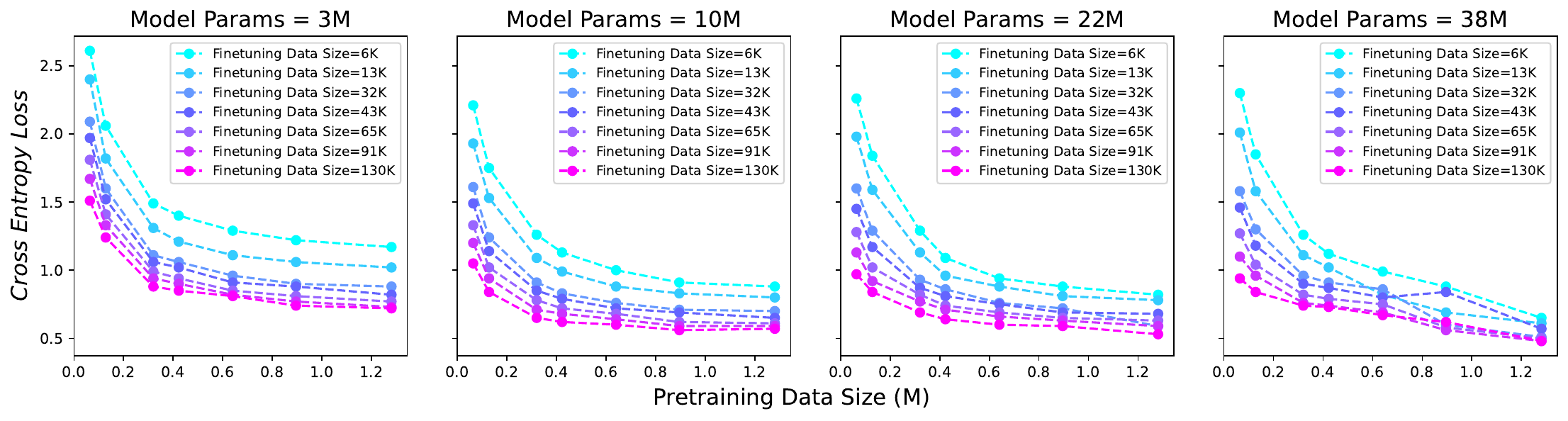}
    \vspace{-10pt}
    \subcaption{Scaling laws for pre-training data size on ImageNet100. Illustration of the impact of pre-training data volume ($Dp$) on cross-entropy loss across different model sizes ($M$) on ImageNet100.}
  \end{minipage}
  \vspace{-10pt}
  \caption{Scaling laws for pre-training data size on ImageNet100. Illustration of the impact of pre-training data volume ($Dp$) on error rate and cross-entropy loss across different model sizes ($M$) on ImageNet100. As $Dp$ increases, both the error rate and cross-entropy loss show a clear downward trend, in line with the power-law scaling behavior proposed in Section \ref{power_law}.}
  \vspace{-5pt}
  \label{upstream_data}
\end{figure*}

% downstream data scaling

\begin{figure*}[h]
  \centering
  \begin{minipage}[b]{0.9\linewidth}
    \includegraphics[width=\linewidth]{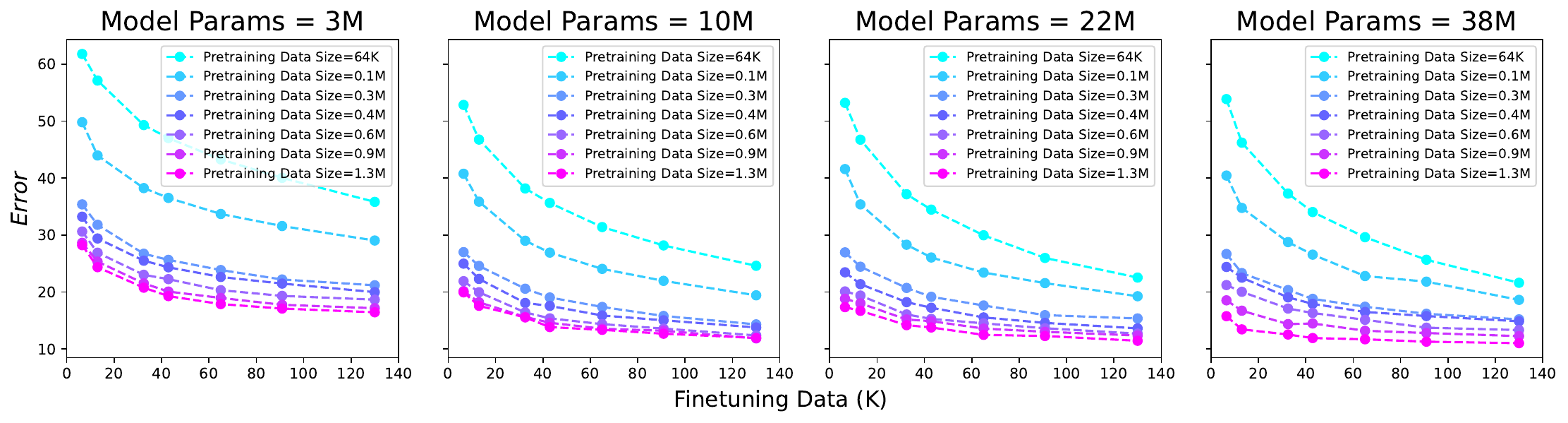}
    \vspace{-10pt}
    \subcaption{Scaling laws for fine-tuning data size on ImageNet100. Illustration of the impact of fine-tuning data volume ($D_f$) on error rate across different model sizes ($M$) on ImageNet100.}
  \end{minipage}
  
\begin{minipage}[b]{0.9\linewidth}
    \includegraphics[width=\linewidth]{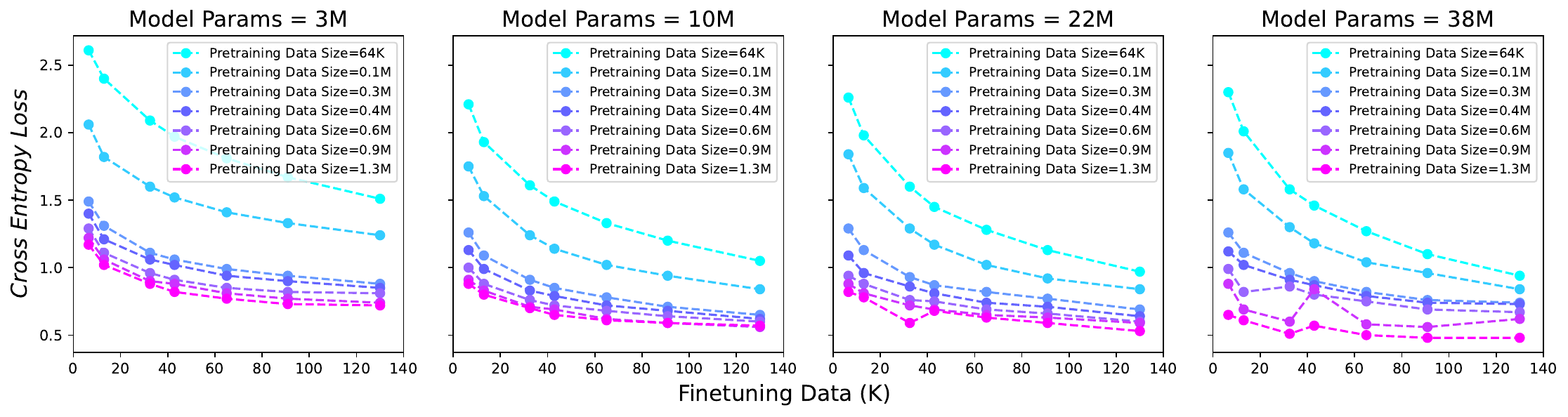}
    \subcaption{Scaling laws for fine-tuning data size on ImageNet100. Illustration of the impact of fine-tuning data volume ($D_f$) on cross-entropy loss across different model sizes ($M$) on ImageNet100.}
    \vspace{-5pt}
  \end{minipage}
  \caption{Scaling laws for fine-tuning data size on ImageNet100. Illustration of the impact of fine-tuning data volume ($D_f$) on error rate and cross-entropy loss across different model sizes ($M$) on ImageNet100. As $Df$ increases, both the error rate and cross-entropy loss show a clear downward trend, in line with the power-law scaling behavior proposed in Section \ref{power_law}.}
  \label{downstream_data}
\end{figure*}

\subsection{Model Distillation: Optimization Strategy}
Knowledge distillation as a small-to-large model guidance technique, a common strategy in large model initialization, offers a potential solution for model compression and knowledge transfer \cite{kd1,kd2,kd3}. By having smaller models guide the training of larger architectures, distillation has the potential to enhance model performance without increasing data volume\cite{kd4,kd5}. However, the relationship between distillation effectiveness and data scale and model size has not been systematically investigated, particularly in data-constrained downstream tasks.

Based on our understanding of scaling laws for downstream tasks, a natural question emerges: How does the effectiveness of knowledge distillation vary with pretraining data scale, model size, and fine-tuning data volume? Is there a critical threshold where the efficacy of distillation strategies fundamentally changes?

\vspace{2pt}
\noindent
\textbf{Small Models Guidance for Larger Models.} To explore these questions, we establish a systematic approach to investigate how models with fewer attention heads can effectively guide the training of larger architectures. This "small-to-large" model framework, inspired by \cite{wang2023data}, leverages the computational efficiency of smaller models while enhancing the performance of larger ones. In our setup, we design experiments with models having 2, 4, 6, and 8 attention heads, conducting three independent distillation processes: distilling from the 2-head to the 4-head model, from 2-head and 4-head models to the 6-head model, and from 2-head, 4-head, and 6-head models to the 8-head model.

To achieve effective knowledge transfer, we implemented a comprehensive distillation loss function design [15], consisting of two core components:

\begin{enumerate}
    \item \textbf{Feature Alignment Loss}: By minimizing the KL divergence between small and large models in the intermediate representation space, we promote consistency in feature representations. Specifically, we apply temperature-scaled soft target alignment to the attention maps at each layer, ensuring the larger model learns critical attention patterns from the smaller model.

    \item \textbf{Output Layer Distillation Loss}: While maintaining the ability to learn from original task labels, we employed a weighted combination loss:
\end{enumerate}
\begin{equation}
L = \alpha \cdot L_{CE}(y_{pred}, y_{true}) + (1-\alpha) \cdot \tau^2 \cdot KL(\sigma(z_s/\tau), \sigma(z_t/\tau))
\end{equation}

where $L_{CE}$ represents the cross-entropy loss, $y_{pred}$ is the model's prediction, and $y_{true}$ is the ground truth label; the KL term represents the Kullback-Leibler divergence between the output distributions of the small and large models, $\sigma$ is the softmax function, $z_s$ and $z_t$ are the logits from the teacher (smaller) and student (larger) models respectively, $\tau$ is the temperature parameter used to smooth the probability distributions, and $\alpha$ is a weight coefficient balancing original task learning and knowledge distillation.

This hierarchical learning approach and carefully designed loss function enable us to analyze how knowledge effectively transfers across models of increasing complexity and to identify optimal pathways for small-to-large model guidance. However, through preliminary experiments, we observe an intriguing phenomenon: \textbf{the effectiveness of distillation appears to have a non-linear relationship with pre-training data scale. Under certain data regimes, distillation strategies excel; in others, traditional training methods prove more effective.} This discovery prompts us to investigate the boundary conditions of knowledge distillation, leading to the development of a new theoretical framework.

\subsection{A Scaling Law for Distilled Models}
To summarize our findings, through systematic experimental analysis, we have successfully fitted a scaling law for distilled models on downstream tasks:

\begin{equation}
E(D_p, M_t, D_f, M_s) = E_{\infty} + \frac{D_p^{-\alpha'}}{\lambda_p'} + \frac{M_s^{-\beta'}}{\lambda_s'} + \frac{D_f^{-\gamma'}}{\lambda_f'} + \frac{M_t^{-\eta'}}{\delta'} \label{e_distill}
\end{equation}

Here, $M_s$ represents the scale of the teacher (smaller) model, while $M_t$ denotes the scale of the student (larger) model. This extended scaling law reveals the quantitative relationship of how pretraining data size ($D_p$), teacher model size ($M_s$), fine-tuning data size ($D_f$), and student model size ($M_t$) collectively influence downstream task performance in a distillation framework. Through this law, we can more precisely predict and optimize resource allocation to maximize model performance under computational constraints.

\begin{figure*}[t!]
  \centering
  \begin{minipage}[b]{0.9\linewidth}
    \includegraphics[width=\linewidth]{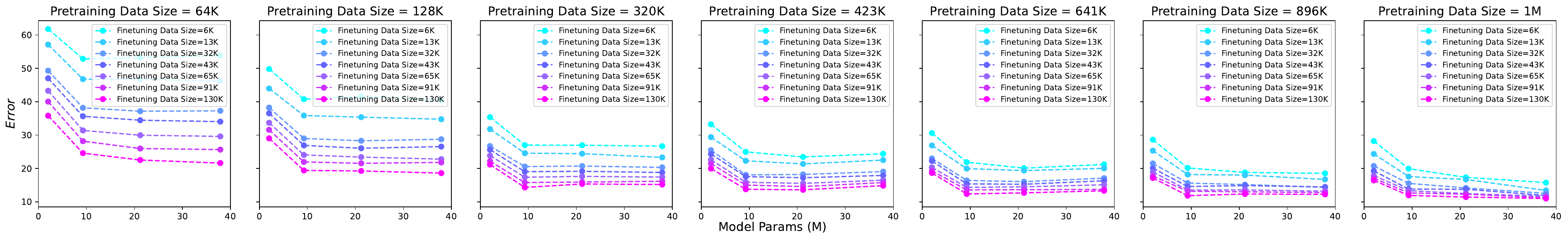}
    \subcaption{Scaling laws for model size on ImageNet100. Illustration of the impact of model size ($M$) on downstream task error rate and cross-entropy loss across different pre-training data sizes ($D_p$) on ImageNet100.}
  \end{minipage}
  
\begin{minipage}[b]{0.9\linewidth}
    \includegraphics[width=\linewidth]{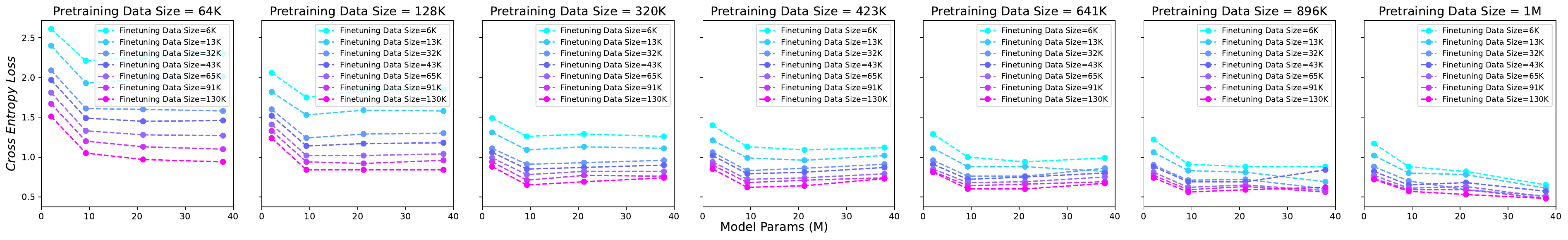}
    \subcaption{Scaling laws for model size on ImageNet100. Illustration of the impact of model size ($M$) on downstream task error rate and cross-entropy loss across different pre-training data sizes ($D_p$) on ImageNet100. }
  \end{minipage}
  \caption{Scaling laws for model size on ImageNet100. Illustration of the impact of model size ($M$) on downstream task error rate and cross-entropy loss across different pre-training data sizes ($D_p$) on ImageNet100. As $M$ increases, both the error rate and cross-entropy loss show a clear downward trend, in line with the power-law scaling behavior proposed in Section \ref{power_law}.}
  \vspace{-5pt}
  \label{downstream_model}
\end{figure*}

\section{Distillation Boundary Theory}
Based on our observations of the relationship between distillation effectiveness and data scale, we propose the Distillation Boundary Theory, which explains the transitions in distillation efficacy under different data conditions. This theory reveals a phase transition in distillation efficiency, providing a theoretical foundation for model training strategy selection.

\begin{theorem}[Distillation Efficacy Boundary]
The efficacy of distilling knowledge from a small to a large model represents not a universal panacea but rather a domain-contingent approach whose utility diminishes beyond a critical threshold of pre-training data availability.
\end{theorem}

\noindent This phenomenon can be formalized through the following mathematical characterization:

\begin{equation}
\left\{
\begin{aligned}
E_1 &= E + \frac{D_p^{-\alpha}}{\lambda_p} + \frac{M^{-\beta}}{\lambda_m} + \frac{D_f^{-\gamma}}{\lambda_f}, \\
E_2 &= E' + \frac{D_p^{-\alpha'}}{\lambda_p'} + \frac{M^{-\beta'}}{\lambda_m'} + \frac{D_f^{-\gamma'}}{\lambda_f'} + \frac{M_t^{-\eta}}{\delta}, \\
\end{aligned}
\right.
\end{equation}

\noindent with the following parametric constraints:

\begin{equation}
\begin{aligned}
E &< E', \quad \gamma > \gamma^\prime, \quad \beta < \beta^\prime, \\
\alpha - \alpha^\prime &\in (-1,0), \quad \lambda_m \approx \lambda_m', \quad \lambda_f \approx \lambda_f'
\end{aligned}
\label{eq:assumptions}
\end{equation}

\noindent such that there exists a critical threshold $D_p^*$:
\begin{equation}
\exists D_p^*, \quad \text{s.t.} \quad E_1(D_p^*) \leq E_2(D_p^*).
\end{equation}

\noindent Here, $E_1$ and $E_2$ contain the error expressions of the standard training paradigm and the small-to-large distillation strategy, respectively. 

\begin{remark}
The conditions \eqref{eq:assumptions}, derived from extensive experimental evidence, reveal fundamental asymmetries in the learning dynamics between traditional and distillation-based approaches. For detailed analysis, please refer to the appendix.
\end{remark}

\subsection{Theoretical Demonstration}
\begin{lemma}[Differential Error Function]
The comparative error differential between the two training paradigms can be represented as a monotonically decreasing function beyond a critical data threshold.
\end{lemma}

\vspace{2pt}
\noindent
\textbf{Proof.} To establish our claim with mathematical rigor, we introduce the comparative error differential function, substituting our error rate formulations yields:

\begin{align}
F(D_p) &= \left(\frac{D_p^{-\alpha}}{\lambda_p} - \frac{D_p^{-\alpha'}}{\lambda_p'}\right) + \left(\frac{M^{-\beta}}{\lambda_m} - \frac{M^{-\beta'}}{\lambda_m'}\right) \nonumber\\
&\quad + \left(\frac{D_f^{-\gamma}}{\lambda_f} - \frac{D_f^{-\gamma'}}{\lambda_f'}\right) - \frac{M_t^{-\eta}}{\delta} + (E - E')
\end{align}

\noindent
We consolidate all terms unrelated to $D_f$ into a constant $\Delta$:
\begin{equation}
\Delta = \left(\frac{M^{-\beta}}{\lambda_m} - \frac{M^{-\beta'}}{\lambda_m'}\right) 
 + \left(\frac{D_f^{-\gamma}}{\lambda_f} - \frac{D_f^{-\gamma'}}{\lambda_f'}\right) - \frac{M_t^{-\eta}}{\delta} + (E - E')
 \label{equation:delta}
\end{equation}
\noindent 
For the first item:
\begin{align}
\frac{M^{-\beta}}{\lambda_m} - \frac{M^{-\beta'}}{\lambda_m'} 
&\approx \frac{1}{\lambda_m} \left( M^{-\beta} - M^{-\beta'} \right) = \frac{M^{-\beta'}}{\lambda_m} \left( M^{\beta' - \beta} - 1 \right)
\label{approx_simplify}
\end{align}
% \begin{align}
% \frac{M^{-\beta}}{\lambda_m} - \frac{M^{-\beta'}}{\lambda_m'} 
% &\approx \frac{1}{\lambda_m} \left( M^{-\beta} - M^{-\beta'} \right) = \frac{M^{-\beta'}}{\lambda_m} \left( M^{\beta' - \beta} - 1 \right)
% \label{approx_simplify}
% \end{align}

\noindent
Given that $M$ represents the model size and is typically a large number (on the order of thousands or more), and that $\beta'$ is empirically observed to be relatively large, the exponential decay term $M^{-\beta'}$ becomes vanishingly small. Consequently, the entire expression in Equation~\eqref{approx_simplify} can be safely regarded as negligible in practice.

\noindent
Similarly, for the second term:
\begin{align}
\frac{D_f^{-\gamma}}{\lambda_f} - \frac{D_f^{-\gamma'}}{\lambda_f'} 
&\approx \frac{1}{\lambda_f} \left( D_f^{-\gamma} - D_f^{-\gamma'} \right) < 0
\label{eq:approx_simplify_df}
\end{align}

\noindent
Based on the analysis above, and considering that the remaining two terms in the expression for \eqref{equation:delta} are also inherently negative, we can approximate that:
\[
\Delta < 0
\]

\noindent The function then simplifies to:
\begin{equation}
F(D_p) = \Delta + \frac{D_p^{-\alpha}}{\lambda_p} - \frac{D_p^{-\alpha'}}{\lambda_p'}
\end{equation}

\noindent
Computing the derivative:
\begin{equation}
F^\prime(D_p) = \frac{\alpha' \cdot D_p^{\alpha-\alpha'} / \lambda_p' - \alpha / \lambda_p}{D_p^{\alpha + 1}}
\end{equation}

\noindent
When $\alpha' \cdot D_p^{\alpha-\alpha'} / \lambda_p' - \alpha / \lambda_p = 0$, we have $F^\prime(D_p) = 0$, yielding the critical point:
\begin{equation}
D_p^* = \left(\frac{\alpha / \lambda_p}{\alpha' / \lambda_p'}\right)^{\frac{1}{\alpha'-\alpha}}
\end{equation}
Since $\alpha - \alpha' \in (-1,0)$, the term $\alpha' \cdot D_p^{-\alpha'} / \lambda_p' - \alpha \cdot D_p^{-\alpha} / \lambda_p$ is monotonically decreasing. We can verify that $F^\prime(D_p)$ is positive when $D_p < D_p^*$ and negative when $D_p > D_p^*$. Therefore, $D_p^*$ is a local maximum of $F(D_p)$. As $D_p \to \infty$, we can derive that $\lim\limits_{D_p \to \infty} F(D_p) = \Delta < 0$.

\noindent
Considering the continuity of $F(D_p)$, and the fact that it is positive for small $D_p$ while $\lim\limits_{D_p \to \infty} F(D_p) < 0$, by the Intermediate Value Theorem, there must exist a point $D_p^{**}$ such that $F(D_p^{**}) = 0$, i.e.:
\begin{equation}
E_1(D_p^{**}) = E_2(D_p^{**})
\end{equation}

\noindent
For all $D_p > D_p^{**}$, we have $F(D_p) < 0$, meaning $E_1(D_p) < E_2(D_p)$, which indicates that traditional training methods outperform distillation methods in this region.

\begin{corollary}
The optimal training strategy selection depends critically on the volume of available pre-training data relative to the derived threshold $D_p^{**}$.
\end{corollary}

\begin{remark}
This theoretical framework provides a principled approach to training strategy selection based on data availability, model architecture, and computational constraints.
\end{remark}

\section{Experiment and Analysis}

\subsection{Experimental Settings}

\vspace{2pt}
\textbf{Model architecture.} We adopt the mainstream Transformer-based DeiT\cite{touvron2021training} architecture as the foundation for our experiments. DeiT performs well with less data and computational resources through optimized training strategies and distillation tokens. We choose DeiT because it represents the current best practices for vision Transformers and is widely adopted across various visual classification tasks.

\vspace{2pt}
\noindent
\textbf{Pre-training dataset.} We use ImageNet-100 as our pre-training dataset, which contains approximately 1.2 million training images distributed across 1000 categories.

\vspace{2pt}
\noindent
\textbf{Downstream dataset.} To evaluate the universality of our scaling laws, we perform fine-tuning on four downstream datasets:
\begin{itemize}
\item{\textbf{TinyImageNet}}:  200 categories with 500 training images per class, 64×64 pixels. This dataset provides a compact yet challenging classification task with smaller image resolutions compared to standard ImageNet.
\item{\textbf{ImageNet100}}: A subset of 100 categories selected from ImageNet1K. This dataset maintains the original image count per class and resolution, offering a medium-complexity transfer learning target.
\item{\textbf{CIFAR100} \cite{cifar}}: 100 categories with 500 training images per class, 32×32 pixels. With its lower resolution and domain shift from ImageNet, CIFAR-100 helps assess cross-domain transfer capabilities.
\item{\textbf{CIFAR10} \cite{cifar}}: 10 categories with 5000 training images per class, 32×32 pixels. This simpler classification task allows us to evaluate scaling behavior on less complex problems with more examples per class.
\end{itemize}

These diverse downstream experiments enable us to comprehensively study the relationships between pre-trained model size, pre-training data scale, and downstream fine-tuning data volume across varying levels of task complexity and domain similarity.

\begin{table}[!t]
\centering
\caption{Fitted exponents for scaling laws on downstream tasks across different datasets.}
\vspace{-10pt}
\label{tab:scaling_coefficients}
\renewcommand{\arraystretch}{1.2}
\resizebox{0.45\textwidth}{!}{
\begin{tabular}{lcccccc}
\toprule
\multirow{2}{*}{\textbf{Dataset}} & \multicolumn{3}{c}{\textbf{Error Rate}} & \multicolumn{3}{c}{\textbf{Cross-Entropy Loss}} \\
\cmidrule(lr){2-4} \cmidrule(lr){5-7}
& $\alpha$ & $\beta$ & $\gamma$ & $\alpha''$ & $\beta''$ & $\gamma''$ \\
\midrule
ImageNet100 & 0.620 & 4.882 & 0.377 & 0.597 & 0.854 & 0.443\\
TinyImageNet & 0.412 & 5.086 & 0.359 & 0.433 & 0.815 & 0.438 \\
CIFAR100 & 0.609 & 1.797 & 0.587 & 0.592 & 0.865 & 0.603 \\
CIFAR10 & 10.129 & 4.975 & 0.331 & 0.667 & 25.435 & 0.391 \\
\bottomrule
\end{tabular}
}
\end{table}

\begin{table}[!t]
\centering
\caption{Fitted exponents for distilled model scaling laws.}
\vspace{-10pt}
\label{tab:scaling_coefficients2}
\renewcommand{\arraystretch}{1.1}

\resizebox{0.4\textwidth}{!}{
\begin{tabular}{lccccc}
\toprule
\multirow{2}{*}{\textbf{Dataset}} & \multicolumn{5}{c}{\textbf{Distilled Model Exponents}} \\
\cmidrule(lr){2-5}
& $\alpha'$ & $\beta'$ & $\gamma'$ & $\eta'$ \\
\midrule
ImageNet100 & 0.702 & 5.840 & 0.338 & 2.053 \\
TinyImageNet & 0.475 & 5.496 & 0.321 & 1.982 \\
\bottomrule
\end{tabular}
}
\end{table}

\begin{figure*}[t!]
  \centering
  \begin{minipage}[b]{0.9\linewidth}
    \includegraphics[width=\linewidth]{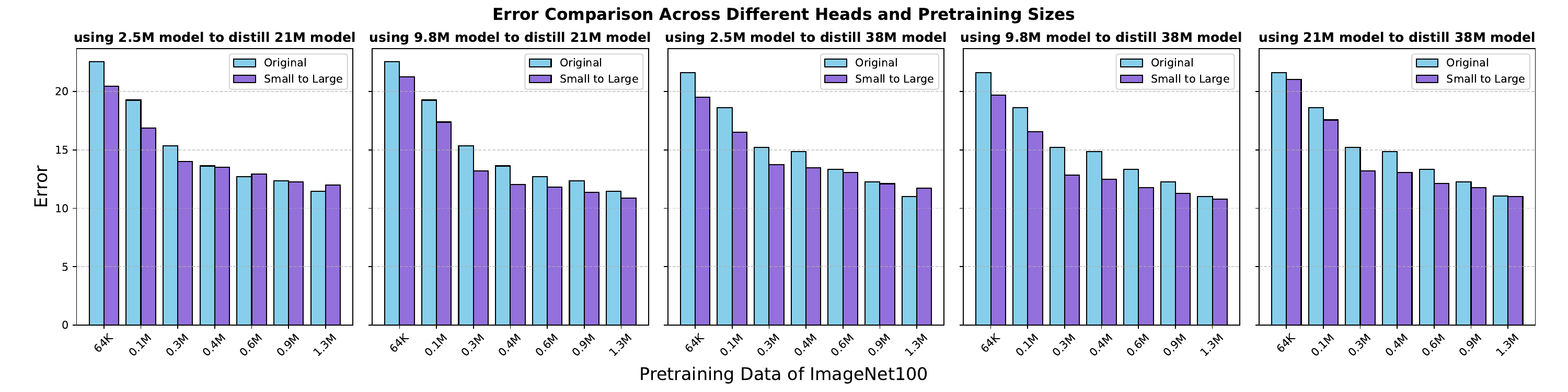}
    % \subcaption{Error comparison between the origin model and distilled approaches on ImageNet100 across varying pretraining data sizes.}
  \end{minipage}
\vspace{-10pt}
  \caption{Evidence for the distillation boundary theory. Error rates of original (blue) and distilled (purple) models are compared across increasing pretraining data sizes (64K–1.3M). With limited data, distillation outperforms the original model; as data grows, the gap narrows and eventually reverses—supporting our theoretical prediction.}
  \vspace{-5pt}
  \label{error_comp}
\end{figure*}

\begin{figure}[h!]
    \centering
    \begin{minipage}[b]{1\linewidth}
    \includegraphics[width=1.0\textwidth]{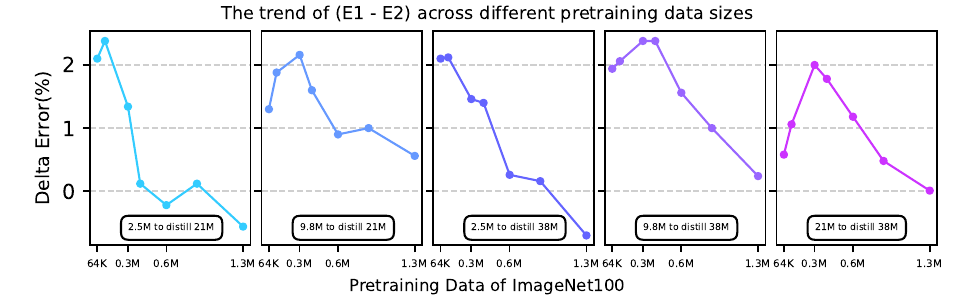} % 
    \subcaption{Delta error between distilled model and original model across different pretraining data sizes on ImageNet100.}
    \end{minipage}

    \begin{minipage}[b]{1\linewidth}
    \includegraphics[width=1.0\textwidth]{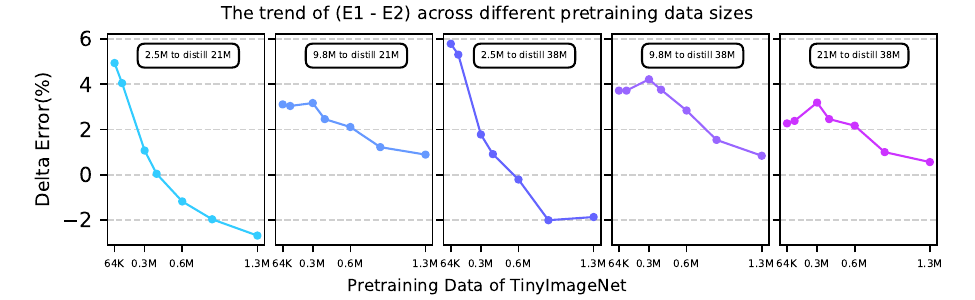} %
    \subcaption{Delta error between distilled model and original model across different pretraining data sizes on TinyImageNet.}
    \end{minipage}
    \vspace{-5pt}
    \caption{Evidence for the distillation boundary theory. As pretraining data size grows, the delta error between distilled and original models first increases and then decreases or steadily decreases, which is consistent with our theoretical analysis.}
    \vspace{-10pt}
    \label{fig:imagenet100_delta}
\end{figure}

\subsection{Main Results of Downstream Scaling Laws}

\vspace{2pt}
\noindent
\textbf{Obs 1. Power law holds for upstream data scaling.} Figure \ref{upstream_data} illustrates and verifies the scaling law of error rate (Equation \ref{e_error}). As the amount of pretraining data increases, the error rate gradually approaches the theoretical minimum value \(E_{\infty}\). The curves for different model scales exhibit similar shapes, but their starting points and descent speeds vary, which may be related to the number of model parameters (\(M\)) and the amount of fine-tuning data (\(D_f\)). Specifically, larger model sizes and more fine-tuning data contribute to reducing the error rate, but the increase in pretraining data remains the key factor for performance improvement. Similarly, for the scaling law of cross-entropy loss (Equation \ref{e_loss}), the figure shows that as \(D_p\) increases, the cross-entropy loss also gradually decreases and approaches the theoretical minimum value \(L_{\infty}\). The loss curves for different model scales exhibit similar characteristics, further confirming the applicability of the power law scaling in upstream data scaling. As the amount of pretraining data increases, the rate of decrease in error rate and cross-entropy loss gradually diminishes, indicating that after reaching a certain data volume, the effect of further increasing the data amount on performance improvement weakens.

\vspace{2pt}
\noindent
\textbf{Obs 2. Power law holds for downstream data scaling.}  Figure \ref{downstream_data} verifies the scaling law of error rate (Equation \ref{e_error}). As the amount of \(D_f\) increases, the error rate gradually decreases and approaches the theoretical minimum value \(E_{\infty}\). For the scaling law of cross-entropy loss (Equation \ref{e_loss}), as \(D_f\) increases, the loss also gradually decreases and approaches the theoretical minimum value \(L_{\infty}\).

\vspace{2pt}
\noindent
\textbf{Obs 3. Power law holds for downstream model scaling.} Figure \ref{downstream_model} verifies the scaling law of error rate (Equation \ref{e_error}). As \(M\) increases, the error rate gradually decreases and approaches the theoretical minimum value \(E_{\infty}\). For the scaling law of cross-entropy loss (Equation \ref{e_loss}), as the \(M\) increases, the loss also gradually decreases and approaches the theoretical minimum value \(L_{\infty}\).

\subsection{Fitting Power Law Exponents}
Table~\ref{tab:scaling_coefficients} presents the fitted exponents for scaling laws across four different datasets. For both error rate and cross-entropy loss functions, we fitted power-law exponents for pretraining data volume ($\alpha$/$\alpha''$), model size ($\beta$/$\beta''$), and fine-tuning data volume ($\gamma$/$\gamma''$). The results demonstrate clear power-law relationships across all datasets, though with notable variations between them. Particularly noteworthy is CIFAR10's unusually strong dependence on pretraining data scaling ($\alpha=10.129$) and model size scaling for loss ($\beta''=25.435$), while other datasets exhibit more moderate exponent values. This phenomenon may stem from CIFAR10's unique characteristics—its low resolution images (32×32 pixels) and limited number of classes (10)—causing more dramatic performance jumps when benefiting from large-scale pretraining. As model size increases, it can more effectively adapt to CIFAR10's specific feature distribution, resulting in steeper scaling curves. ImageNet100 and TinyImageNet show similar behavior in model size scaling for error rate ($\beta$ of $4.882$ and $-5.086$ respectively), potentially reflecting their comparable complexity. Overall, these fitted parameters provide a solid foundation for understanding and predicting model performance under various conditions.

Table~\ref{tab:scaling_coefficients2} shows the fitted exponents for distilled model scaling laws across two datasets. We fitted power-law exponents for pretraining data volume ($\alpha'$), model size ($\beta'$), fine-tuning data volume ($\gamma'$), and teacher model size ($\eta'$). ImageNet100 and TinyImageNet show opposite signs for pretraining data and model size parameters, indicating different responses to these factors. However, both exhibit similar positive dependence on teacher model size ($\eta' \approx 2$) and comparable fine-tuning data exponents (0.338 and 0.321). These findings support our theoretical analysis of the distillation boundary effect, confirming that as the pre-training data improves, knowledge transfer efficiency undergoes significant changes.

\subsection{Small Model Guidance: Scaling Optimization}
We conduct comprehensive experiments on our proposed small-to-large model transfer method across four downstream tasks of varying complexity: CIFAR10, CIFAR100, TinyImageNet, and ImageNet100. Figure ~\ref{error_comp} in the main text presents our detailed experimental results on ImageNet100, with complete results for all other datasets available in the Appendix. 

Our experimental results demonstrate that employing the small-to-large distillation approach leads to improved model performance and smoother loss curves, particularly under limited finetuning data conditions. However, through both empirical evidence and theoretical analysis, we establish the existence of a critical finetuning data threshold beyond which the small-to-large training strategy becomes less effective than the conventional pretrain-and-finetune paradigm. This phenomenon can also be intuitively understood: a small model, constrained by its limited capacity, cannot fully encapsulate the extensive knowledge inherent in large-scale data. Consequently, excessive guidance from a small model trained on an overwhelming amount of data is more likely to introduce biases, causing the larger model to internalize incorrect information. This ultimately results in suboptimal performance compared to training the large model independently. These findings hold practical implications, highlighting that even when employing knowledge distillation, careful consideration of dataset size and model capacity is essential to achieving optimal performance.

\section{Conclusion}
This paper establishes systematic scaling laws for visual downstream experiment under data-constrained conditions, not only revealing stable power-law relationships between model size, pretraining data, and fine-tuning data volumes but also proposing the distillation boundary theory through theoretical and empirical investigation. This theory clearly defines the optimal application scenarios for small-to-large model knowledge transfer, identifying a critical transition point where distilled models shift from initial performance advantages to being surpassed by traditional methods as fine-tuning data increases. Unlike existing research primarily focused on large-scale pretraining scenarios, our work directly addresses the common challenge of data scarcity in practical applications, providing researchers with theoretical guidance for optimizing model selection and training strategies in resource-limited environments. This enables more precise performance prediction and more efficient resource allocation in visual learning tasks where data collection is costly, thereby filling a critical theoretical gap in the field of visual transfer learning.
% \begin{acks}
% To Robert, for the bagels and explaining CMYK and color spaces.
% \end{acks}

%%
%% The next two lines define the bibliography style to be used, and
%% the bibliography file.

\bibliographystyle{ACM-Reference-Format}
\bibliography{sample-base}

\appendix
\twocolumn[
  \begin{center}
    \Huge\bfseries Scaling Laws for Data-Efficient Visual Transfer Learning
    \vspace{0.5cm} % 可选，添加一些空间
  \end{center}
]
\section{Experimental Settings}
\textbf{Hyper-parameter configuration.} 
Table \ref{tab:training_params} outlines the training parameter configurations used in our experiments. For the pretraining phase, we employed the ImageNet dataset with an AdamW optimizer and a batch size of 128. To ensure training stability, we set the base learning rate to 5e-4 with a warmup learning rate of 1e-6. For downstream task fine-tuning, we adopted dataset-specific configurations: larger datasets (TinyImageNet and ImageNet100) maintained the same optimizer and batch size as pretraining but with an adjusted warmup learning rate of 1e-5 to facilitate better adaptation; for smaller datasets (CIFAR100 and CIFAR10), we reduced the batch size to 16, which helps improve model performance under data-limited conditions. These parameter configurations were designed to validate the effectiveness of our proposed scaling laws in data-constrained environments.

\begin{table}[t!]
\centering
\caption{Experimental Parameters for Training.}
\vspace{-10pt}
\renewcommand{\arraystretch}{1.2}
\resizebox{0.45\textwidth}{!}{
\begin{tabular}{lccccc}
\toprule
\textbf{Task} & \textbf{Dataset} & \textbf{Optimizer} & \textbf{Batch Size} & \textbf{LR} & \textbf{Warmup LR} \\
\midrule
\multirow{1}{*}{Pretraining} & ImageNet & AdamW & 128 & 5e-4 & 1e-6 \\
\midrule
\multirow{4}{*}{Downstream} & TinyImageNet & AdamW & 128 & 5e-4 & 1e-5 \\
 & ImageNet100 & AdamW & 128 & 5e-4 & 1e-5 \\
 & CIFAR100 & AdamW & 16 & 5e-4 & 1e-5 \\
 & CIFAR10 & AdamW & 16 & 5e-4 & 1e-5 \\
\bottomrule
\end{tabular}}
\label{tab:training_params}
\end{table}

\section{Results on Different Upstream Datasets}
\begin{figure*}[h]
  \centering
  \begin{minipage}[b]{\linewidth}
    \includegraphics[width=\linewidth]{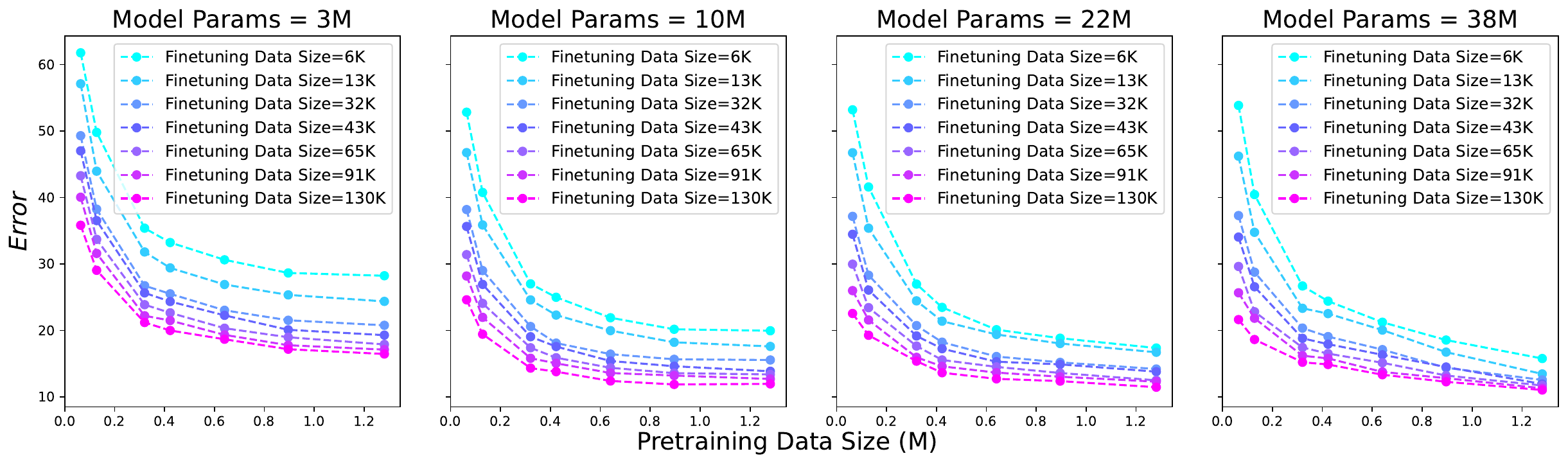}
  \end{minipage}
  
\begin{minipage}[b]{\linewidth}
    \includegraphics[width=\linewidth]{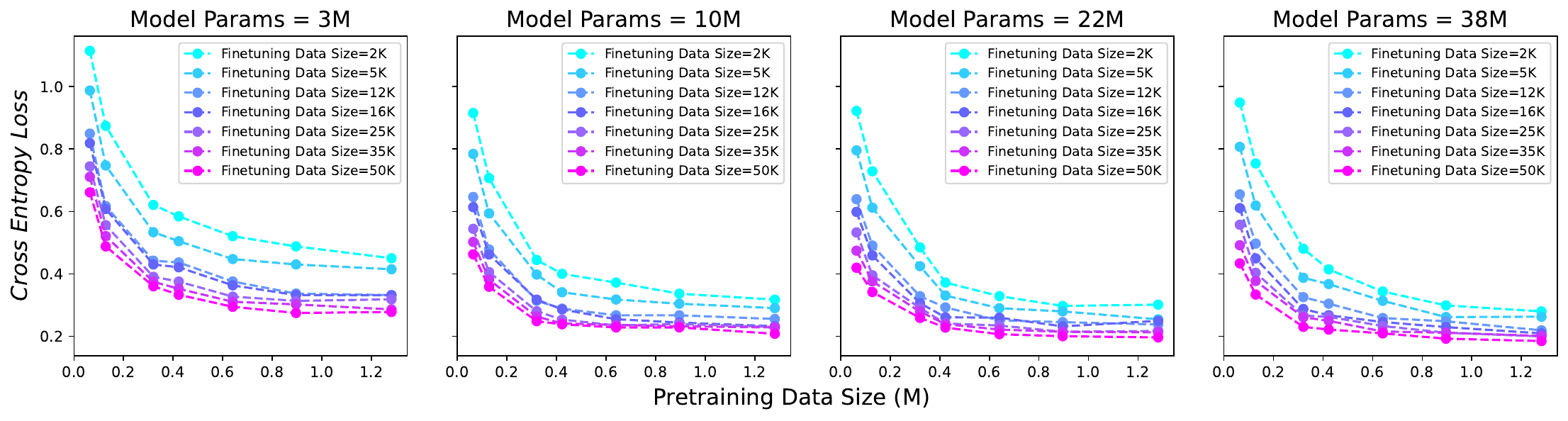}
  \end{minipage}
   \setlength{\abovecaptionskip}{-5pt}
  \setlength{\belowcaptionskip}{-5pt}
  \caption{Impact of pre-training data volume ($D_p$) on error rate and loss across different model sizes ($M$) on CIFAR10.}
  \label{upstream_data:cifar10}
\end{figure*}

\begin{figure*}[h]
  \centering
  \begin{minipage}[b]{\linewidth}
    \includegraphics[width=\linewidth]{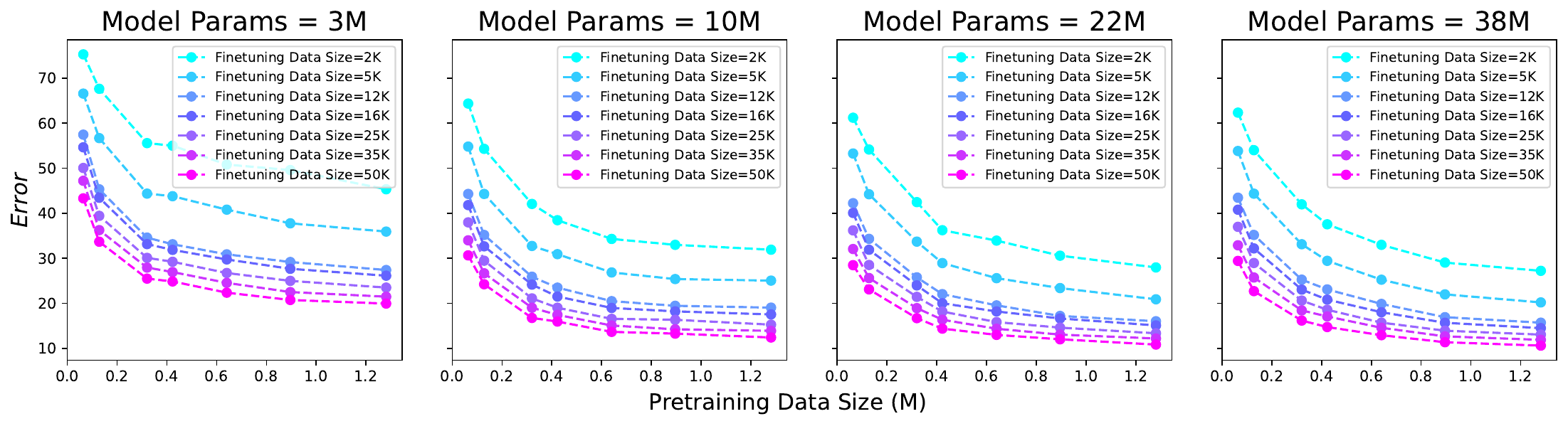}
  \end{minipage}
  
  \begin{minipage}[b]{\linewidth}
    \includegraphics[width=\linewidth]{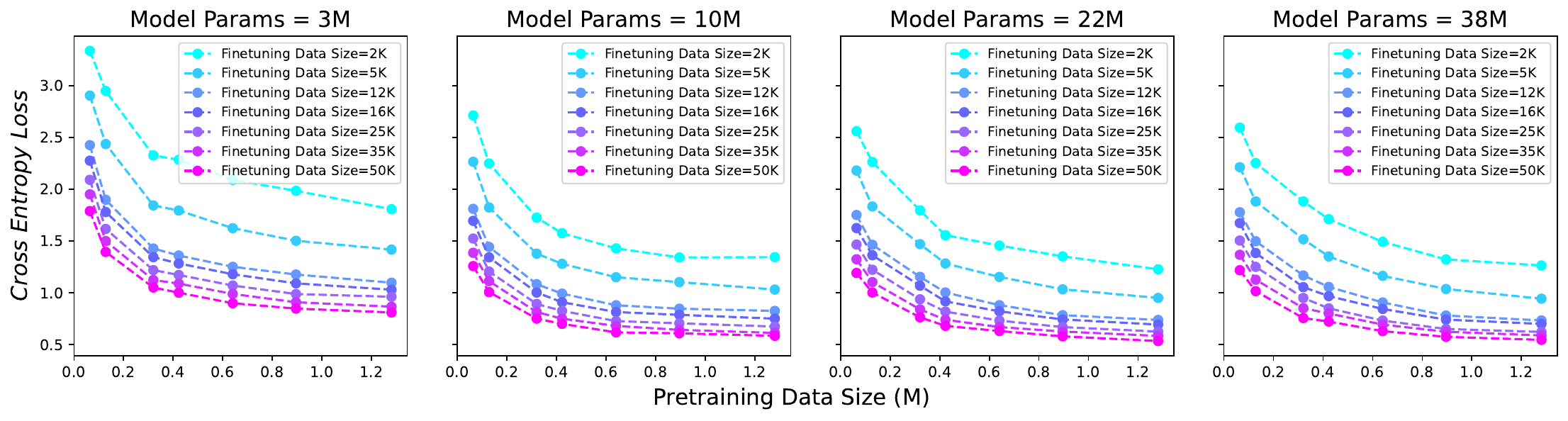}
  \end{minipage}
   \setlength{\abovecaptionskip}{-5pt}
  \setlength{\belowcaptionskip}{-5pt}
  \caption{Impact of pre-training data volume ($D_p$) on error rate and loss across different model sizes ($M$) on CIFAR100.}
  \label{upstream_data:cifar100}
\end{figure*}

\begin{figure*}[h]
  \centering
  \begin{minipage}[b]{\linewidth}
    \includegraphics[width=\linewidth]{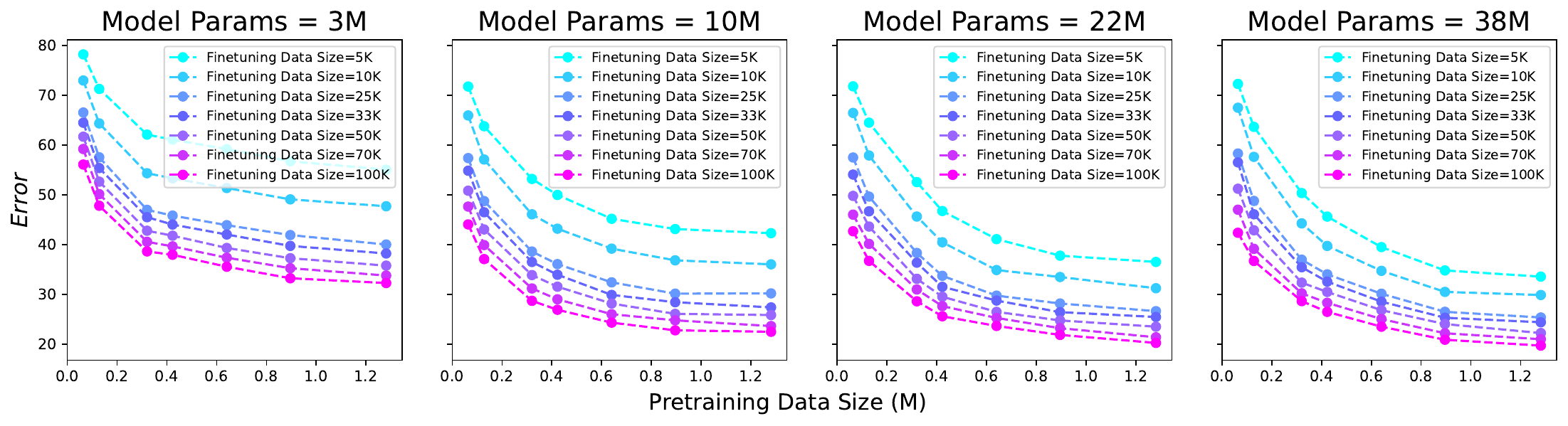}
  \end{minipage}
  
  \begin{minipage}[b]{\linewidth}
        \includegraphics[width=\linewidth]{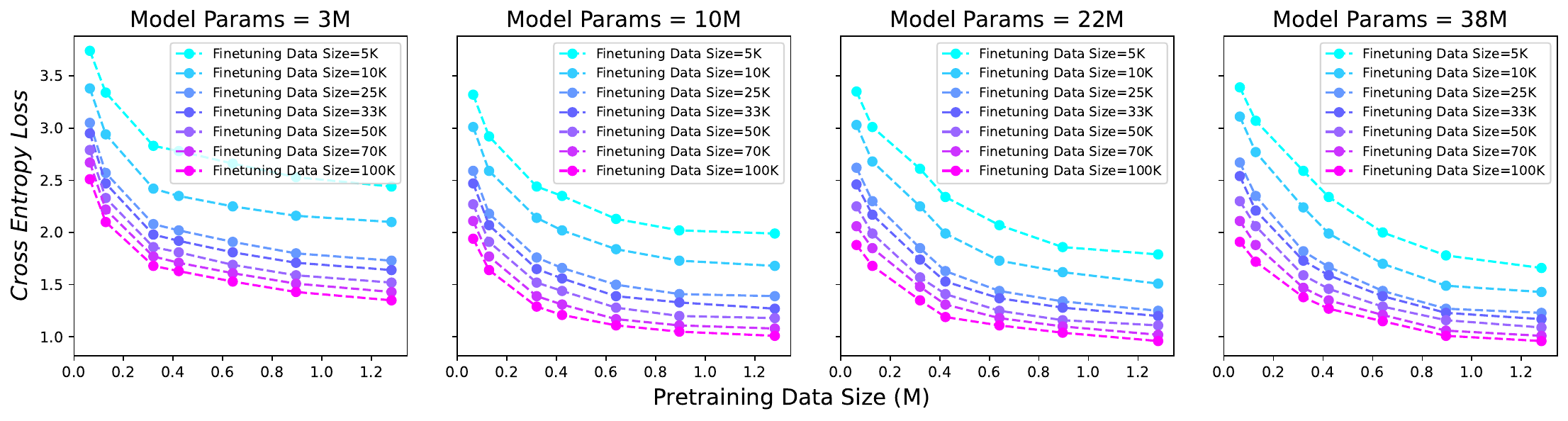}
  \end{minipage}
  \setlength{\abovecaptionskip}{-5pt}
  \setlength{\belowcaptionskip}{-5pt}
    \caption{Impact of pre-training data volume ($D_p$) on error rate and loss across different model sizes ($M$) on TinyImageNet.}
  \label{upstream_data:tiny}
\end{figure*}

 Fig. \ref{upstream_data:cifar10}, \ref{upstream_data:cifar100}, \ref{upstream_data:tiny} illustrate the impact of pre-training data volume on the error rate and loss in downstream tasks across the CIFAR-10, CIFAR-100, and TinyImageNet datasets. As the pre-training data volume increases, both the error rate and cross-entropy loss exhibit a clear downward trend, closely resembling the curve patterns observed for the ImageNet dataset presented in the main text.

\section{Results on Different Downstream Datasets}
\begin{figure*}[h]
  \centering
  \begin{minipage}[b]{\linewidth}
    \includegraphics[width=\linewidth]{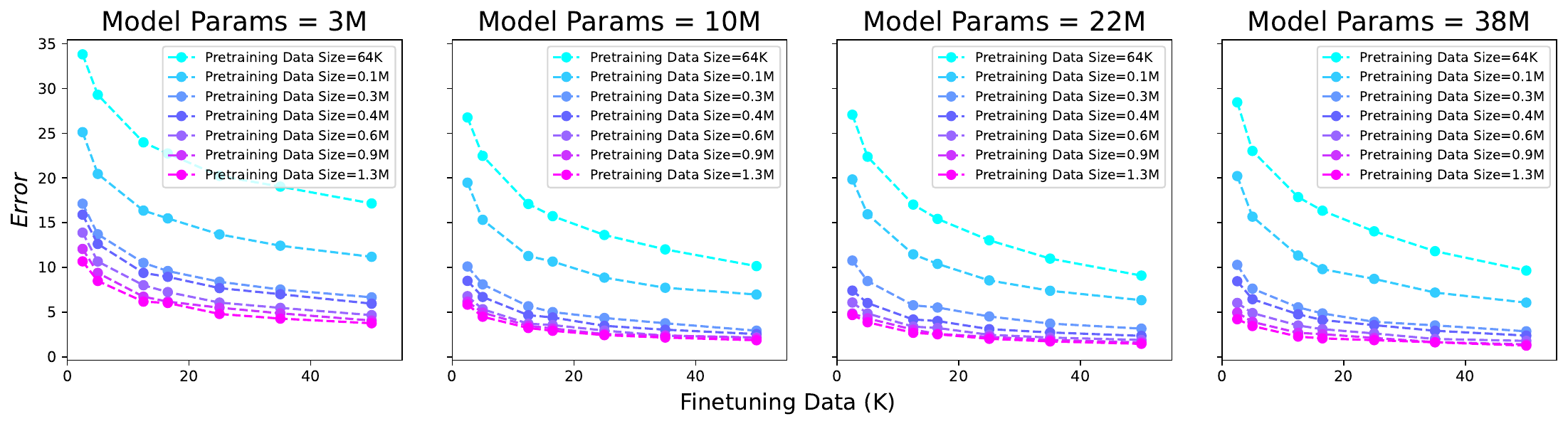}
  \end{minipage}
  
\begin{minipage}[b]{\linewidth}
    \includegraphics[width=\linewidth]{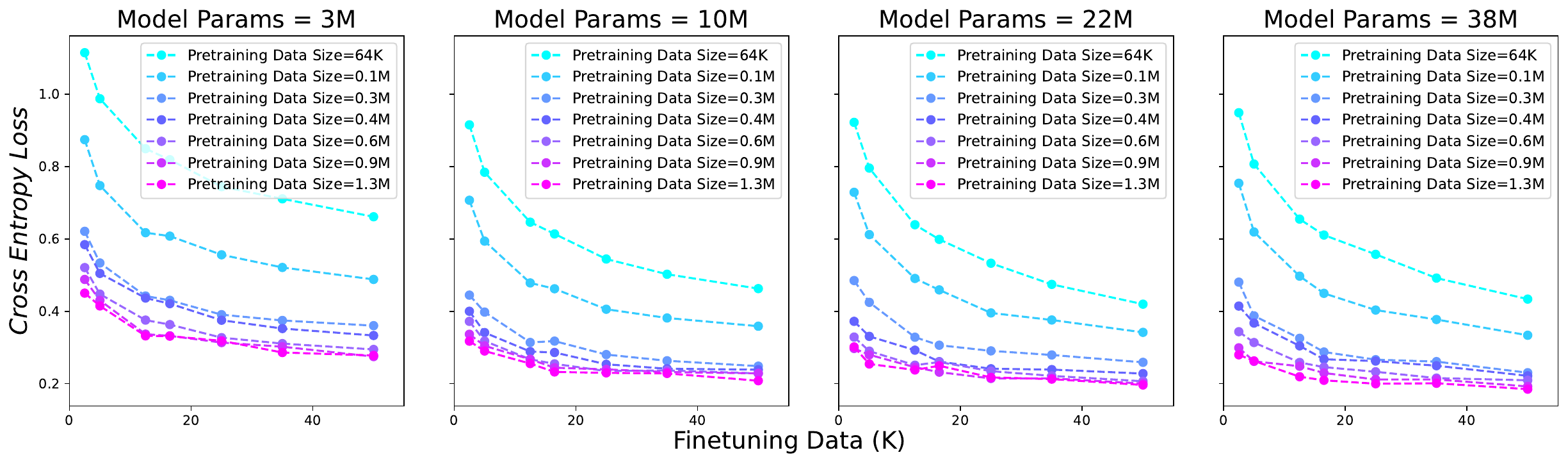}
  \end{minipage}
   \setlength{\abovecaptionskip}{-5pt}
  \setlength{\belowcaptionskip}{-5pt}
  \caption{Impact of finetuning data volume ($D_f$) on error rate and loss across different model sizes ($M$) on CIFAR10.}
  \label{downstream_data:cifar10}
\end{figure*}

\begin{figure*}[h]
  \centering
  \begin{minipage}[b]{\linewidth}
    \includegraphics[width=\linewidth]{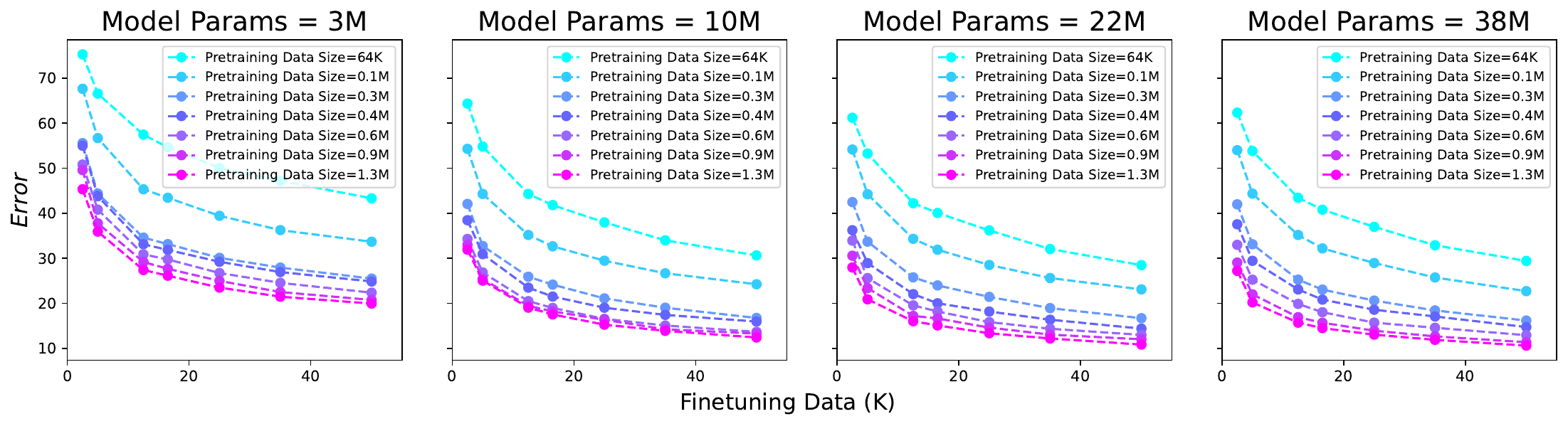}
  \end{minipage}
  
  \begin{minipage}[b]{\linewidth}
        \includegraphics[width=\linewidth]{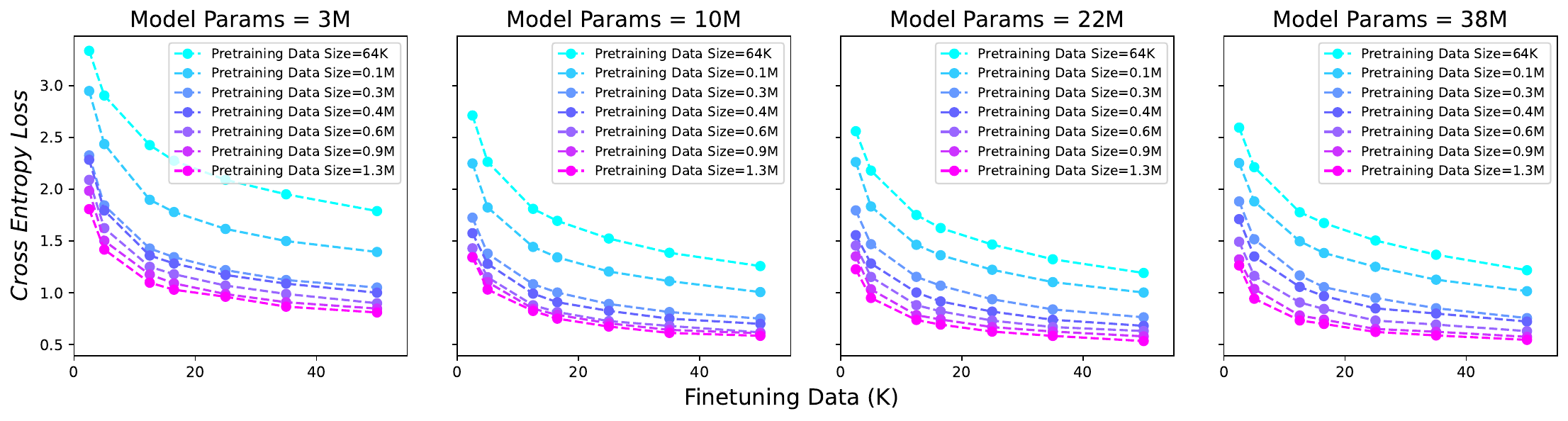}
  \end{minipage}
   \setlength{\abovecaptionskip}{-5pt}
  \setlength{\belowcaptionskip}{-5pt}
  \caption{Impact of finetuning data volume ($D_f$) on error rate and loss across different model sizes ($M$) on CIFAR100.}
  \label{downstream_data:cifar100}
\end{figure*}

\begin{figure*}[h]
  \centering
  \begin{minipage}[b]{\linewidth}
    \includegraphics[width=\linewidth]{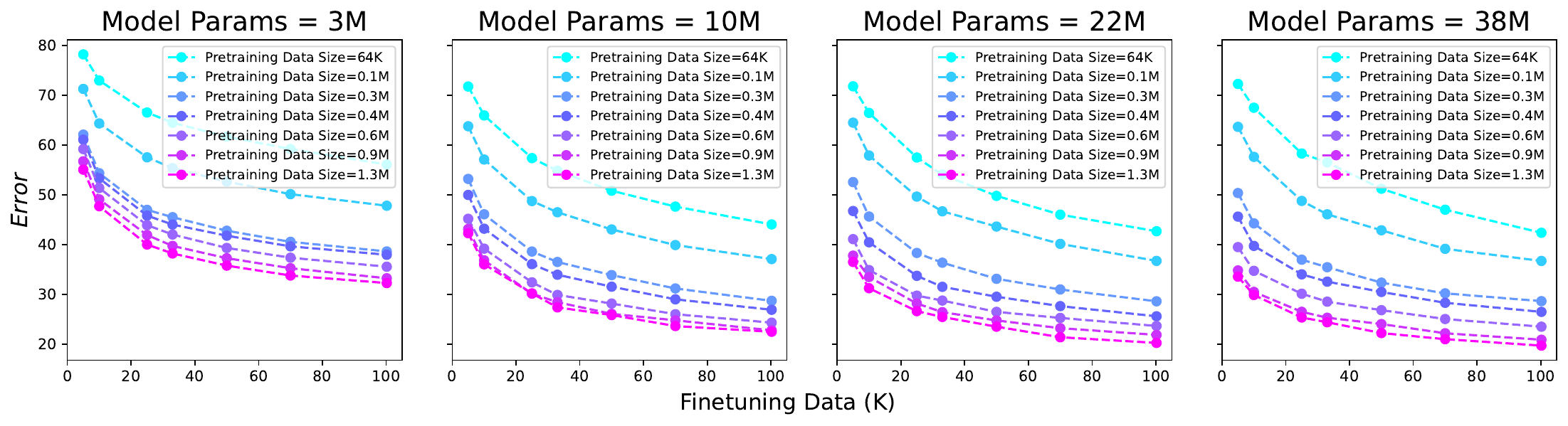}
  \end{minipage}
  
  \begin{minipage}[b]{\linewidth}
        \includegraphics[width=\linewidth]{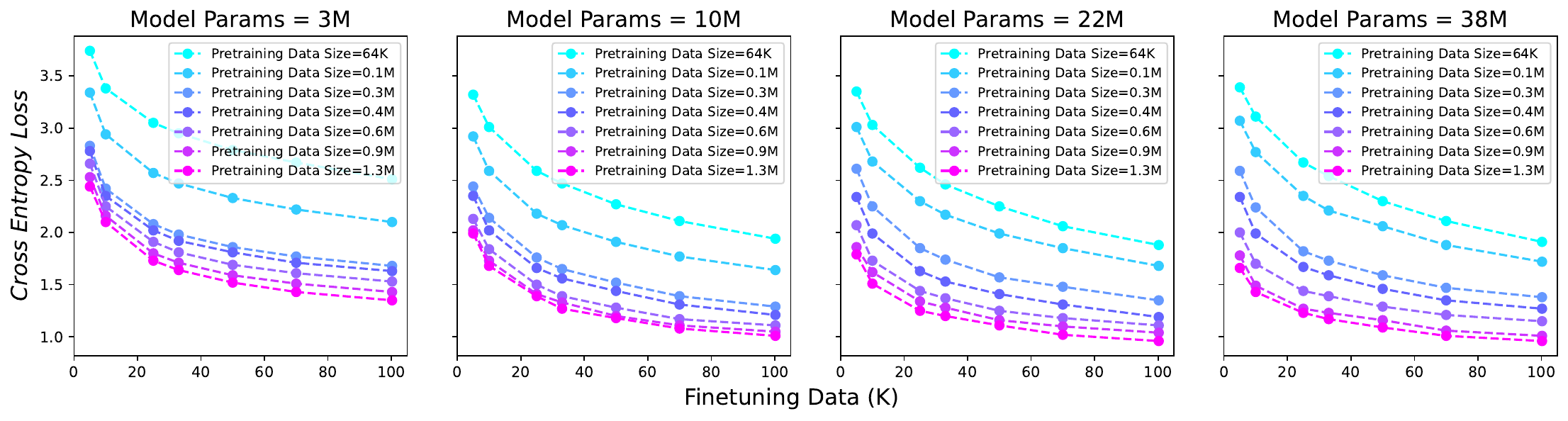}
  \end{minipage}
  \setlength{\abovecaptionskip}{-5pt}
  \setlength{\belowcaptionskip}{-5pt}
    \caption{Impact of finetuning data volume ($D_f$) on error rate and loss across different model sizes ($M$) on TinyImageNet.}
  \label{downstream_data:tiny}
\end{figure*}

 As shown in Fig.\ref{downstream_data:cifar10}, \ref{downstream_data:cifar100}, \ref{downstream_data:tiny}, the impact of downstream data on error rate and loss in downstream tasks across CIFAR-10, CIFAR-100, and TinyImageNet. The curves consistently exhibit a power-law pattern, aligning well with the functional form of the equation presented in the main text.

\section{Results on Different Downstream Models}
\begin{figure*}[h]
  \centering
  \begin{minipage}[b]{\linewidth}
    \includegraphics[width=\linewidth]{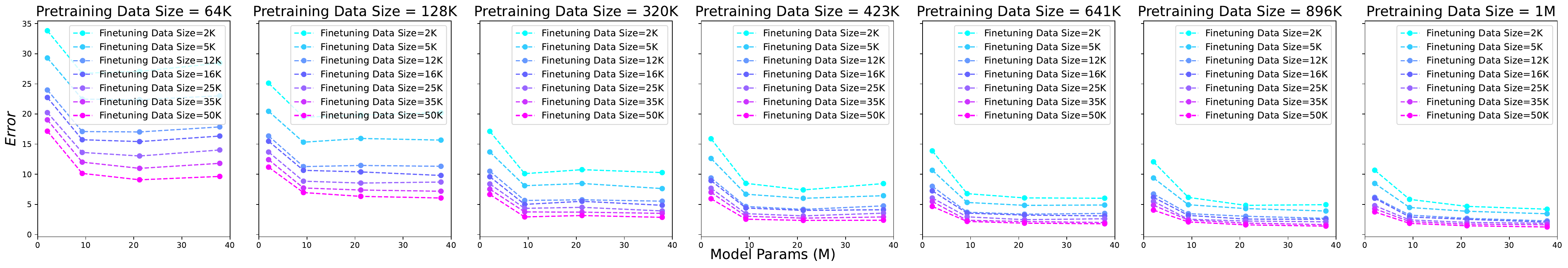}
  \end{minipage}
  
  \begin{minipage}[b]{\linewidth}
    \includegraphics[width=\linewidth]{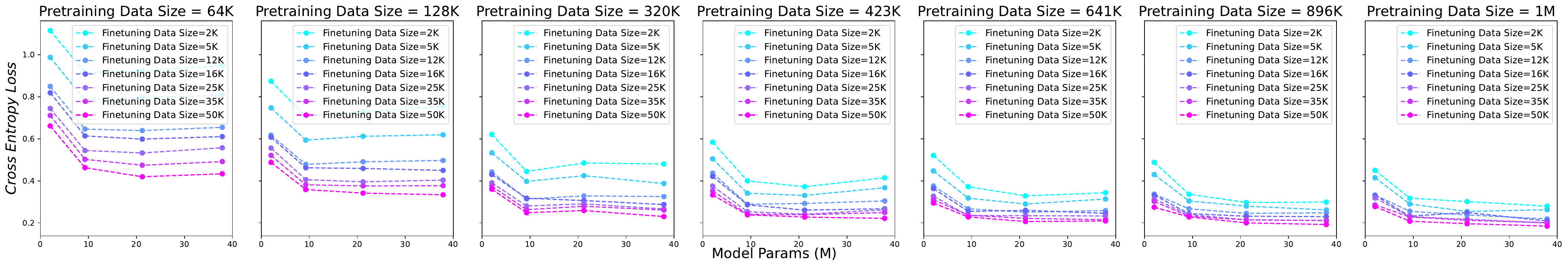}
  \end{minipage}
  \setlength{\abovecaptionskip}{-7pt}
  \setlength{\belowcaptionskip}{-7pt}
  \caption{Impact of model size ($M$) on downstream task error rate and cross-entropy loss across different pre-training data sizes ($D_p$) on CIFAR10.}
  \label{downstream_model:cifar10}
\end{figure*}

\begin{figure*}[h]
  \centering
  \begin{minipage}[b]{\linewidth}
    \includegraphics[width=\linewidth]{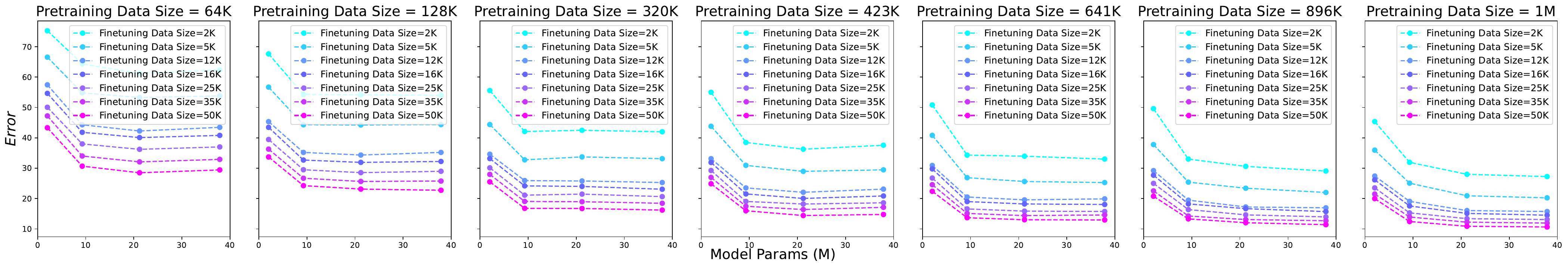}
  \end{minipage}
  
  \begin{minipage}[b]{\linewidth}
    \includegraphics[width=\linewidth]{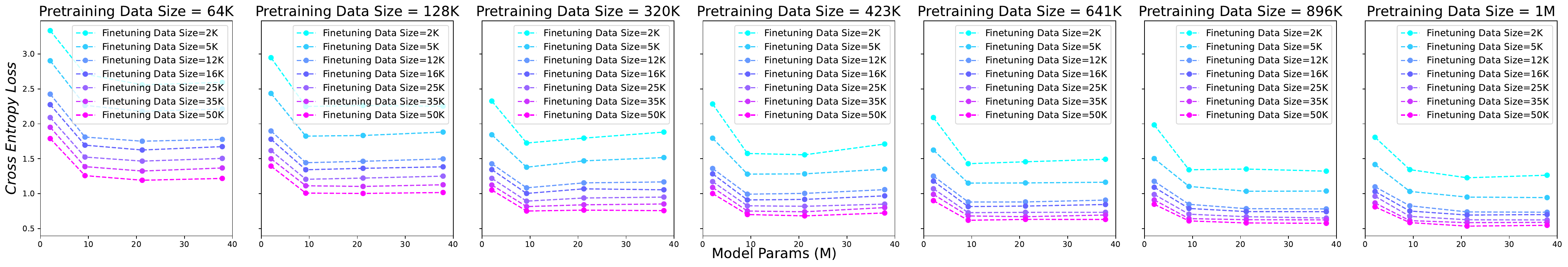}
  \end{minipage}
  \setlength{\abovecaptionskip}{-7pt}
  \setlength{\belowcaptionskip}{-7pt}
  \caption{Impact of model size ($M$) on downstream task error rate and cross-entropy loss across different pre-training data sizes ($D_p$) on CIFAR100.}
  \label{downstream_model:cifar100}
\end{figure*}

\begin{figure*}[h]
  \centering
  \begin{minipage}[b]{\linewidth}
    \includegraphics[width=\linewidth]{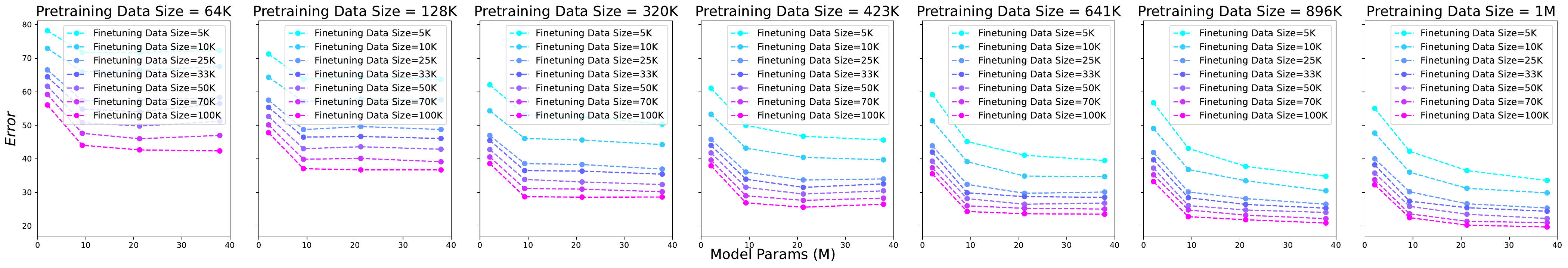}
  \end{minipage}
  
  \begin{minipage}[b]{\linewidth}
    \includegraphics[width=\linewidth]{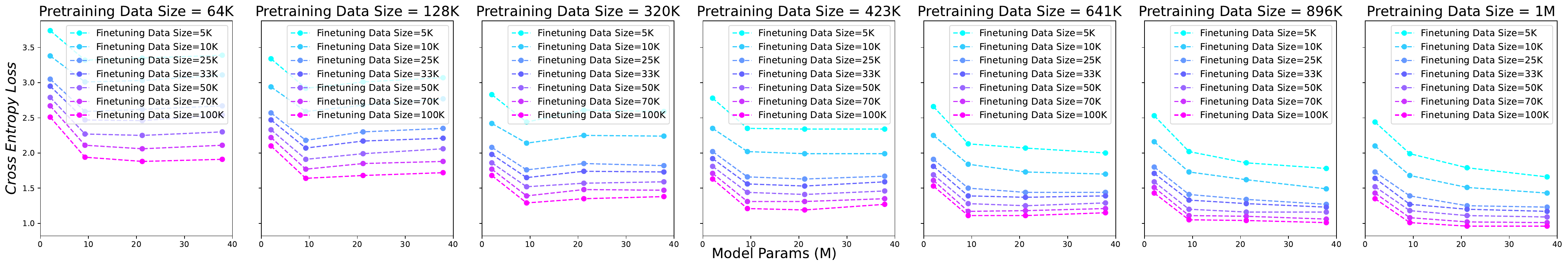}
  \end{minipage}
  \setlength{\abovecaptionskip}{-7pt}
  \setlength{\belowcaptionskip}{-7pt}
  \caption{Impact of model size ($M$) on downstream task error rate and cross-entropy loss across different pre-training data sizes ($D_p$) on TinyImageNet.}
  \label{downstream_model:tiny}
\end{figure*}

 From the Fig. \ref{downstream_model:cifar10},\ref{downstream_model:cifar100},\ref{downstream_model:tiny} we can also observe that as the downstream model size increases, the error rate and loss do not simply decrease; in some cases, they even increase with model size. This further highlights the importance of exploring the relationship between performance, parameter count, and data volume in visual transfer learning.

\section{Results of Fitted Coefficients}
Table~\ref{tab:scaling_coefficients_sup} presents the fitted coefficients for scaling laws across various downstream computer vision tasks. Our analysis encompasses four prominent datasets: ImageNet100, TinyImageNet, CIFAR100, and CIFAR10. For each dataset, we report coefficients for both error rate and cross-entropy loss scaling behaviors.

The error rate coefficients reveal interesting patterns across datasets. The asymptotic error rate $E_\infty$ approaches extremely small values in all cases, with TinyImageNet showing the most dramatic potential for improvement ($1.76 \times 10^{-27}$). The parameter scaling exponent $\lambda_p$ is notably highest for CIFAR10 (1.84), suggesting that this dataset benefits substantially more from increased model size compared to others. Similarly, CIFAR10 exhibits the strongest model scaling coefficient $\lambda_m$ at 1.22, while CIFAR100 shows minimal benefits from model scaling ($2.60 \times 10^{-6}$). Data scaling effects, captured by $\lambda_f$, are most pronounced for ImageNet100 (0.179).

For cross-entropy loss, the asymptotic values $L_\infty$ are more consistent across ImageNet100, TinyImageNet, and CIFAR100 (all approximately $10^{-5}$), while CIFAR10 demonstrates a remarkably lower potential limit ($1.38 \times 10^{-13}$). The parameter scaling coefficients $\lambda_p$ for loss are relatively uniform across datasets, suggesting similar benefits from increased parameterization regardless of the specific visual classification task. The model and data scaling coefficients ($\lambda_m$ and $\lambda_f$) show larger variations, indicating dataset-specific scaling properties.

\begin{table}[h!]
    \centering
    \caption{Fitted coefficients for scaling laws on downstream tasks across different datasets.}
    \vspace{-10pt}
    \label{tab:scaling_coefficients_sup}
    \renewcommand{\arraystretch}{1.2}
    
    % ------------------- Error Rate -------------------
    \begin{tabular}{lcccc}
    \toprule
    \multirow{2}{*}{\textbf{Dataset}} & \multicolumn{4}{c}{\textbf{Error Rate}}\\
    & $E_\infty$ & $\lambda_p$ & $\lambda_m$ & $\lambda_f$ \\
    \midrule
    ImageNet100     &1.44e-14   & 4.39e-3 & 3.05e-2 & 1.79e-1 \\
    TinyImageNet    &1.76e-27   & 2.58e-2 & 4.55e-1 & 1.44e-1 \\
    CIFAR100        &2.55e-15   & 3.83e-3 & 2.60e-6 & 3.45e-2 \\
    CIFAR10         &2.17e-10   & 1.84e+0 & 1.22e+0 & 5.53e-1 \\
    \bottomrule
    \end{tabular}
    % ------------------- Cross-Entropy Loss -------------------
    \begin{tabular}{lcccc}
    \toprule
    \multirow{2}{*}{\textbf{Dataset}} & \multicolumn{4}{c}{\textbf{Cross Entropy Loss }}\\
    & $L_\infty$ & $\lambda_p$ & $\lambda_m$ & $\lambda_f$ \\
    \midrule
    ImageNet100     &1.98e-05   & 1.18e-3 & 1.15e-5 & 2.58e-2 \\
    TinyImageNet    &1.39e-05   & 4.72e-3 & 1.3e-5  & 1.62e-2 \\
    CIFAR100        &2.68e-05   & 1.14e-3 & 6.25e-6 & 7.30e-3 \\
    CIFAR10         &1.38e-13  & 1.23e-3 & 1.01e-2 & 1.37e-1 \\
    \bottomrule
    \end{tabular}
    \vspace{-10pt}
\end{table}

\section{Limitations}
\begin{itemize}
  % \item Our study is limited to the vision modeling setting. Although extensive empirical evidence suggests that the functional form of scaling laws holds across various domains, we cannot be absolutely certain that the scaling behavior of transfer learning aligns with our findings in all domains.
  \item Our empirical findings are derived from experiments conducted on four benchmark datasets. Though these datasets are widely used, they may not fully capture the heterogeneity and complexity of the broader visual domain. 
  \item Compared to the NLP domain, scaling law research in computer vision remains relatively underdeveloped. One primary reason is the significant disparity in both the scale of datasets and the size of models typically used in vision compared to NLP—often differing by several orders of magnitude. This gap limits the direct transfer of scaling insights from NLP to vision.
\end{itemize}

\end{document}